\documentclass[runningheads]{llncs}
\usepackage{graphicx}
\usepackage{comment}
\usepackage{booktabs}
\usepackage{amsmath,amssymb} 
\usepackage{color}

\usepackage{boldline}
\usepackage{lipsum}
\usepackage{fixltx2e}
\usepackage{float}
\usepackage{amsmath}
\usepackage{amsfonts}
\usepackage{amssymb}
\usepackage{bm}
\usepackage{bbm}
\usepackage{mathtools}
\usepackage{enumitem}
\usepackage{thmtools,thm-restate}
\usepackage[ruled,vlined]{algorithm2e}

\def\AA{\mathcal{A}}\def\BB{\mathcal{B}}\def\CC{\mathcal{C}}
\def\DD{\mathcal{D}}\def\EE{\mathcal{E}}\def\FF{\mathcal{F}}

\def\LL{\mathcal{L}}
\def\OO{\mathcal{O}}

\def\SSX{\mathcal{S}}\def\TT{\mathcal{T}}
\def\XX{\mathcal{X}}
\def\YY{\mathcal{Y}}

\def\bb{\mathbf{b}}
\def\eb{\mathbf{e}}

\def\ob{\mathbf{o}}
\def\rb{\mathbf{r}}

\def\wb{\mathbf{w}}
\def\yb{\mathbf{y}}\def\zb{\mathbf{z}}

\def\Rbb{\mathbb{R}}

\def\R{\Rbb}

\DeclareMathOperator*{\argmin}{argmin}
\DeclareMathOperator*{\argmax}{argmax}

\newcommand{\bpsi}{\boldsymbol{\psi}}

\usepackage{xspace}
\makeatletter
\DeclareRobustCommand\onedot{\futurelet\@let@token\@onedot}
\def\@onedot{\ifx\@let@token.\else.\null\fi\xspace}

\def\ie{\emph{i.e}\onedot}

\def\etal{\emph{et al}\onedot}
\makeatother

\usepackage{etex,etoolbox}
\usepackage{environ}

\makeatletter

\usepackage{hyperref}
\usepackage[capitalise]{cleveref}
\usepackage{cite}

\usepackage{multirow}
\usepackage{lipsum}

\begin{document}
\pagestyle{headings}
\mainmatter
\def\ECCVSubNumber{4628}  

\newcommand{\amirreza}[1]{{\leavevmode\color{blue} Amirreza: #1}}
\newcommand{\amir}[1]{{\leavevmode\color{blue} Amir: #1}}
\newcommand{\AJ}[1]{{\leavevmode\color{magenta} AJ: #1}}

\newcommand\blfootnote[1]{%
  \begingroup
  \renewcommand\thefootnote{}\footnote{#1}%
  \addtocounter{footnote}{-1}%
  \endgroup
}

\title{Pairwise Similarity Knowledge Transfer for Weakly Supervised Object Localization}

\titlerunning{Pairwise Similarity Knowledge Transfer for WSOL}
%
\author{Amir Rahimi$^\star$\inst{1} \and Amirreza Shaban$^\star$\inst{2} \and Thalaiyasingam Ajanthan\inst{1} \and Richard Hartley\inst{1,3} \and Byron Boots\inst{4}}
%
\authorrunning{A.~Rahimi, A.~Shaban~\etal}
%
\institute{\textsuperscript{1}ANU, ACRV ,  
\textsuperscript{2} Georgia Tech, 
\textsuperscript{3} Google Research, 
\textsuperscript{4}University of Washington \\
Corresponding authors: \email{amir.rahimi@anu.edu.au}, \email{ashaban@uw.edu}.\\
$\text{ }^\star${Authors contributed equally.}}
\maketitle
\begin{abstract}
\vspace{-0.5cm}
Weakly Supervised Object Localization (WSOL) methods 
only require image level labels as opposed to expensive bounding box annotations required by fully supervised algorithms. We study the problem of learning localization model on target classes with weakly supervised image labels, helped by a fully annotated source dataset.
Typically, a WSOL model is first trained to predict class generic objectness scores on an off-the-shelf fully supervised source dataset and then it is progressively adapted to learn the objects in the weakly supervised target dataset.
In this work, we argue that learning only an objectness function is a weak form of knowledge transfer and propose to learn a classwise pairwise similarity function that directly compares two input proposals as well.
The combined localization model and the estimated object annotations are jointly learned in an alternating optimization paradigm as is typically done in standard WSOL methods.
In contrast to the existing work that learns pairwise similarities, our approach optimizes a unified objective with convergence guarantee and it is computationally efficient for large-scale applications.
Experiments on the COCO and ILSVRC $2013$ detection datasets show that the performance of the localization model improves significantly with the inclusion of pairwise similarity function. For instance, in the ILSVRC dataset, the Correct Localization (CorLoc) performance improves from $72.8\%$ to $78.2\%$ which is a new state-of-the-art for WSOL task in the context of knowledge transfer.
\end{abstract}
\section{Introduction}
    Weakly Supervised Object Localization (WSOL) methods have gained a lot of attention in computer vision~\cite{wan2019c,gao2019note,arun2019dissimilarity,uijlings2018revisiting,cinbis2016weakly,bilen2014weakly,deselaers2010localizing,tang2017multiple,tang2018pcl}. Despite their supervised counterparts~\cite{he2017mask,redmon2016you,liu2016ssd,singh2018sniper,lin2017focal} that require the object class and their bounding box annotations, WSOL methods only require the image level labels indicating presence or absence of  object classes. In spite of major improvements~\cite{wan2019c,cinbis2016weakly} in this area of research, there is still a large performance gap between weakly supervised and fully supervised object localization algorithms. In a successful attempt, WSOL methods are adopted to use an already annotated object detection dataset, called source dataset, to improve the weakly supervised learning performance in new classes~\cite{uijlings2018revisiting,hoffman2016large}. These approaches learn transferable knowledge from the source dataset and use it to speed up learning new categories in the weakly supervised setting. \\
\indent Multiple Instance Learning (MIL) methods like MI-SVM~\cite{andrews2003support} are the predominant methods in weakly supervised object localization~\cite{wan2019c,cinbis2016weakly,bilen2014weakly}. Typically, images are decomposed into bags of object proposals and the problem is posed as selecting one proposal from each bag that contains an object class. MIL methods take advantage of alternating optimization to progressively learn a classwise objectness (unary) function and the optimal selection in re-training and re-localization steps, respectively. Typically, the source dataset is used to learn an initial generic objectness function which is used to steer the selection toward objects and away from background proposals~\cite{uijlings2018revisiting,hoffman2016large,bilen2016weakly,rochan2015weakly,tang2014co,guillaumin2012large}. However, solely learning an objectness measure is a sub-optimal form of knowledge transfer as it can only discriminate objects from background proposals, while it is unable to discriminate between different object classes. Deselaers~\etal~\cite{deselaers2010localizing} propose to additionally learn a pairwise similarity function from the fully annotated dataset and frame WOSL as a graph labeling problem where nodes represent bags and each proposal corresponds to one label for the corresponding node. The edges which reflect the cost of wrong pairwise labeling are derived from the learned pairwise similarities. Additionally, they propose an ad-hoc algorithm to progressively adapt the scoring functions to learn the weakly supervised classes using alternating re-training and re-localization steps. Unlike the alternating optimization in MIL, re-training and re-localization steps in~\cite{deselaers2010localizing} does not optimize a unified objective and therefore the convergence of their method could not be guaranteed. Despite good performance on medium scale problems, this method is less popular especially in large scale problems where computing all the pairwise similarities is intractable. \\
\indent In this work, we adapt the localization model in MIL to additionally learn a pairwise similarity function and use a two-step alternating optimization to jointly learn the augmented localization model and the optimal selection. 
In the re-training step, the pairwise and unary functions are learned given the current selected proposals for each class. In the re-localization step, the selected proposals are updated given the current pairwise and unary similarity functions. 
We show that with a properly chosen  localization loss function, the objective in the re-localization step can be equivalently expressed as a graph labeling problem very similar to the model in~\cite{deselaers2010localizing}. We use the computationally effective iterated conditional modes (ICM) graph inference algorithm~\cite{besag1986statistical} in the re-localization step which updates the selection of one bag in each iteration. Unfortunately, the ICM algorithm is prone to local minimum and its performance is highly dependent on the quality of its initial conditions. Inspired by the recent work on few-shot object localization~\cite{shaban2019learning}, we divide the dataset into smaller mini-problems and solve each mini-problem individually using TRWS~\cite{kolmogorov2006convergent}. We combine the solutions of these mini-problems to initialize the ICM algorithm. Surprisingly, we observe that initializing ICM with the optimal selection from mini-problems of small sizes considerably improves the convergence point of ICM. 

Our work addresses the main disadvantages of graph labeling algorithm in~\cite{deselaers2010localizing}. First, we formulate learning pairwise and unary functions and updating the optimal proposal selections with graph labeling within a two-step alternating optimization framework where each step is optimizing a unified objective and the convergence is guaranteed. Second, we propose a computationally efficient graph inference algorithm which uses a novel initialization method combined with ICM updates in the re-localization step. Our experiments show our method significantly improves the performance of MIL methods in large-scale COCO~\cite{lin2014microsoft} and ILSVRC 2013 detection~\cite{russakovsky2015imagenet} datasets. 
Particularly, our method sets a new state-of-the-art performance of $78.2\%$ correct localization~\cite{deselaers2010localizing} for the WSOL task in the ILSVRC 2013 detection dataset\footnote{Source code is available on \href{https://github.com/AmirooR/Pairwise-Similarity-knowledge-Transfer-WSOL}{https://github.com/AmirooR/Pairwise-Similarity-knowledge-Transfer-WSOL}}.

\section{Related Work}
We review the MIL based algorithms among other branches in WSOL~\cite{bilen2016weakly,tang2014co}. These approaches exploit alternating optimization to learn a detector and the optimal selection jointly. The algorithm iteratively alternates between re-localizing the objects given the current detector and re-training the detector given the current selection. In the recent years, alternating optimization scheme combined with deep neural networks has been the state-of-the-art in WSOL~\cite{wan2019c,gao2019note,zhu2017soft}. However, due to the non-convexity of its objective function, this method is prone to local minimum which typically leads to sub-optimal results~\cite{bilen2015weakly,wan2018min} e.g. selecting the salient parts instead of the whole object. Addressing this issue has been the main focus of research in WSOL in the recent years~\cite{cinbis2016weakly,kumar2010self,wan2019c}. In multi-fold~\cite{cinbis2016weakly}, weakly supervised dataset is split into separate training and testing folds to avoid overfitting. Kumar~\etal~\cite{kumar2010self} propose an iterative self-paced learning algorithm that gradually learns from easy to hard samples to avoid getting stuck in bad local optimum points. Wan~\etal~\cite{wan2019c} propose a continuation MIL algorithm to smooth out the non-convex loss function in order to alleviate the local optimum problem in a systematic way. \\
\indent Transfer learning is another way to improve WSOL performance. These approaches utilize the information in a fully annotated dataset to learn an improved object detector on a weakly supervised dataset~\cite{uijlings2018revisiting,hoffman2016large,rochan2015weakly,guillaumin2012large}. They leverage the common visual information between object classes to improve the localization performance in the target weakly supervised dataset. In a standard knowledge transfer framework, the fully annotated dataset is used to learn a class agnostic objectness measure. This measure is incorporated during the alternating optimization step to steer the detector toward objects and away from the background~\cite{uijlings2018revisiting}. Although the objectness measure is a powerful metric in differentiating between background and foreground, it fails to discriminate between different object classes. Several works have utilized pairwise similarity measure for improving WSOL~\cite{shaban2019learning,deselaers2010localizing,tang2014co}. Shaban~\etal~\cite{shaban2019learning} use a relation network to predict pairwise similarity between pairs of proposals in the context of few-shot object co-localization. Deselaers~\etal~\cite{deselaers2010localizing} frame WSOL as a graph labeling problem with pairwise and unary potentials and progressively adapt the potential functions to learn weakly supervised classes. Tang~\etal~\cite{tang2014co} utilize the pairwise similarity between proposals to capture the inter-class diversity for the co-localization task. Hayder~\etal~\cite{hayder2014object,hayder2015structural} use pairwise learning for object co-detection.


\section{Problem Description and Background}
We review the standard dataset definition and optimization method for the weakly supervised object localization problem~\cite{wan2019c,uijlings2018revisiting,cinbis2016weakly,deselaers2010localizing}.
\subsubsection{Dataset and Notation.}
Suppose each image is decomposed into a collection of object proposals which form a bag $\BB=\{\eb_i\}_{i=1}^m$ where an object proposal ${\eb_i \in \R^d}$ is represented by a $d$-dimensional feature vector. We denote ${y(\eb) \in \CC \cup \{c_\varnothing\}}$ the label for object proposal $\eb$. In this definition $\CC$ is a set of object classes and $c_\varnothing$ denotes the background class.
Given a class $c\in\CC$ we can also define the binary label
\begin{equation}
y_c(\eb) = \begin{cases}
    1 \text{ if } y(\eb) = c \\
    0 \text{ otherwise.}
\end{cases}
\end{equation}
With this notation a dataset is a set of bags along with the labels. For a weakly supervised dataset, only bag-level labels that denote the presence/absence of objects in a given bag are available. More precisely, the label for bag $\BB$ is written as $\YY(\BB) = \{ c \mid \exists \eb \in \BB \text{ s.t. } y(\eb) = c\in\CC\}$. Let $Y_c(\BB) \in \{0,1\}$ denote the binary bag label which indicates the presence/absence of class $c$ in bag $\BB$. \\
Given a weakly supervised dataset $\DD_\TT = \{\TT, \YY_\TT\}$ called the target dataset, with $\TT=\{\BB_j\}_{j=1}^N$ and corresponding bag labels $\YY_\TT = \{\YY(\BB)\}_{\BB \in \TT}$, 
the goal is to estimate the latent proposal unary labeling\footnote{Notice, the labeling is a function defined over a finite set of variables, which can be treated as a vector. Here, $\yb_c$ denotes the vector of labels $y_c(\eb)$ for all proposals $\eb$.} $\yb_c$ for all object classes $c\in\CC_\TT$ in the target set.\\
For ease of notation, we also introduce a pairwise labeling function between pairs of proposals. The pairwise labeling function $r:\R^d\times\R^d\to\{0,1\}$ is designated to output $1$ when two object proposals belong to the same object class and $0$ otherwise, \ie,
\begin{equation}\label{eq:relation}
r(\eb, \eb') = \begin{cases}
    1 \text{ if } y(\eb) = y(\eb') \neq c_\varnothing\\
    0 \text{ otherwise.}
\end{cases}
\end{equation}
Likewise, given a class $c$, two proposals are related under the class conditional pairwise labeling function ${r_c:\R^d\times\R^d\to\{0,1\}}$ if they both belong to class $c$. Similar to the unary labeling, since the pairwise labeling function is also defined over a finite set of variables, it can be seen as a vector. Unless we use the word vector or function, the context will determine whether we use the unary or pairwise labeling  as a vector or a function. We use the ``hat'' notation to refer to the estimated (pseudo) unary or pairwise labeling by the weakly supervised learning algorithm.
%

\subsubsection{Multiple Instance Learning (MIL).} In standard MIL~\cite{andrews2003support}, the problem is solved by jointly learning a unary score function $\psi^\mathrm{U}_c: \R^d \to \R$ (typically represented by a neural network) and a feasible (pseudo) labeling
$\hat{\yb}_c$ that minimize the empirical unary loss
\begin{equation}
\label{eq:loss_mil}
    \LL^\mathrm{U}_c(\bpsi^\mathrm{U}_c, \hat{\yb}_c\mid\TT) = \sum_{\BB \in \TT} \sum_{\eb \in \BB} \ell(\psi^\mathrm{U}_c(\eb), \hat{y}_c(\eb)),
\end{equation}
where the loss function $\ell:\R\times\{0,1\}\to\R$ measures the incompatibility between predicted scores $\psi^\mathrm{U}_c(\eb)$ and the pseudo labels $\hat{y}_c(\eb)$. 
Here, likewise to the labeling, we denote the class score for all the proposals as a vector $\bpsi^\mathrm{U}_c$.
Note that the unary labeling $\hat{\yb}_c$ is feasible if exactly one proposal has label $1$ in each positive bag, and every other proposal has label $0$~\cite{cinbis2016weakly}. To this end, the set of feasible labeling $\FF$ can be defined as
%
\begin{equation}\label{eq:feasible_label}
\FF = \left\{\hat{\yb}_c\mid \hat{y}_c(\eb) \in\{0,1\}, \sum_{\eb\in \BB} \hat{y}_c(\eb) = Y_c(\BB), \forall \BB \in \TT\right\}.
\end{equation}
%
Finally, the problem is framed as minimizing the loss over all possible vectors $\bpsi^\mathrm{U}_c$ (\ie, unary functions represented by the neural network) and the feasible labels $\hat{\yb}_c$
\begin{equation}\label{eq:rev_opt}
\begin{aligned}
    \min_{\bpsi^\mathrm{U}_c,\hat{\yb}_c} &\LL^\mathrm{U}_c(\bpsi^\mathrm{U}_c, \hat{\yb}_c\mid\TT), \\
    &\text{s.t. } \hat{\yb}_c \in \FF.
\end{aligned}
\end{equation}
%
\subsubsection{Optimization.}
This objective is typically minimized in an iterative two-step alternating optimization paradigm~\cite{ortega1970iterative}. The optimization process starts with some initial value of the parameters and labels, and iteratively alternates between {\em re-training} and {\em re-localization} steps until convergence. In the re-training step, the parameters of the unary score function $\psi^\mathrm{U}_c$ are optimized while the labels $\hat{\yb}_c$ are fixed. In the re-localization step, proposal labels are updated given the current unary scores. The optimization in the re-localization step is equivalent to assigning positive label to the proposal with the highest unary score within each positive bag and label $0$ to all other proposals~\cite{andrews2003support}. Formally, label of the proposal $\eb \in \BB$ in bag $\BB$ is updated as
\begin{equation} \label{eq:mil_reloc}
    \hat{y}_c(\eb) = \begin{cases}
        1 ~\text{if $Y_c(\BB) = 1$ and $\eb = \argmax_{\eb'\in \BB} \psi^\mathrm{U}_c(\eb')$} \\
        0 ~\text{otherwise.}
    \end{cases}
\end{equation}
\subsubsection{Knowledge Transfer.}
In this paper, we also assume having access to an auxiliary fully annotated dataset $\DD_\SSX$ (source dataset) with object classes in $\CC_\SSX$ which is a disjoint set from the target dataset classes, \ie, $\CC_\TT \cap \CC_\SSX=\varnothing$. In the standard practice~\cite{uijlings2018revisiting,rochan2015weakly,guillaumin2012large}, the source dataset is used to learn a class agnostic unary score $\psi^\mathrm{U}:\R^d\to\R$ which measures how likely the input proposal $\eb$ tightly encloses  a foreground object. Then, the unary score vector used in \cref{eq:mil_reloc} is adapted to $\bpsi^\mathrm{U}_c \leftarrow \lambda\bpsi^\mathrm{U}_c + (1-\lambda)\bpsi^\mathrm{U}$ 
for some $0 \leq \lambda \leq 1$. This steers the labeling toward choosing proposals that contain complete objects. Although the class agnostic unary score function $\psi^\mathrm{U}$ is learned on the source classes, since objects share common properties, it transfers to the unseen classes in the target set.

\section{Proposed Method}
In addition to learning the unary scores, we also learn a classwise pairwise similarity function $\psi^\mathrm{P}_c:\R^d\times\R^d\to\R$ that estimates the pairwise labeling between pairs of proposals. That is for the target class $c$, pairwise similarity score $\psi^\mathrm{P}_c(\eb,\eb')$ between two input proposals $\eb,\eb' \in \R^d$ has a high value if two proposals are related, \ie, $\hat{r}_c(\eb,\eb') = 1$ and a low value otherwise. We define the empirical pairwise similarity loss to measure the incompatibility between pairwise similarity function predictions and the pairwise labeling $\hat{\rb}_c$
\begin{equation}
    \LL^\mathrm{P}_c(\bpsi^\mathrm{P}_c, \hat{\rb}_c|\TT) = \sum_{\substack{\BB, \BB' \in \TT \\ \BB \neq \BB'}} \sum_{\substack{\eb \in \BB \\ \eb' \in \BB'}}  \ell(\psi^\mathrm{P}_c(\eb, \eb'),
    \hat{r}_c(\eb, \eb')),
\end{equation}
where $\bpsi^\mathrm{P}_c$ denotes the vector of the pairwise similarities of all pairs of proposals, and ${\ell:\R\times\{0,1\}\to\R}$ is the loss function. We define the overall loss as the weighted sum of the empirical pairwise similarity and the unary loss 
\begin{equation} \label{eq:loss}
    \LL_c(\bpsi_c, \hat{\zb}_c|\TT) = \alpha\LL^\mathrm{P}_c(\bpsi^\mathrm{P}_c, \hat{r}_{c}|\TT) + \LL^\mathrm{U}_c(\bpsi^\mathrm{U}_c, \hat{y}_{c}|\TT),
\end{equation} 
where $\bpsi_c = \left[\bpsi^\mathrm{U}_c, \bpsi^\mathrm{P}_c\right]$ is the vector of unary and pairwise similarity scores combined, and $\hat{\zb}_c = \left[\hat{\yb}_c, \hat{\rb}_c\right]$ denotes the concatenation of unary and pairwise labeling vectors, and $\alpha>0$ controls the importance of the pairwise similarity loss.

We employ alternating optimization to jointly optimize the loss over the parameters of the scoring functions $\psi^\mathrm{U}_c$ and $\psi^\mathrm{P}_c$ (re-training) and labelings $\hat{\zb}_c$ (re-localization). In re-training, the objective function is optimized to learn the pairwise similarity and the unary scoring functions from the pseudo labels. In re-localization, we use the current scores to update the labelings. 

Training the model with fixed labels, \ie re-training step, is straightforward and can be implemented within any common neural network framework. We use sigmoid cross entropy loss in both empirical unary and pairwise similarity losses
\begin{equation}\label{eq:sig_cross_entropy}
\ell(x, y) = -(1-y) \log(1-\sigma(x)) - y \log(\sigma(x)),
\end{equation}
where $x \in \R$ is the predicted logit, $y \in \{0, 1\}$ is the label, and $\sigma:\R\to\R$ denotes the sigmoid function ${\sigma(x)=1/(1+\exp(-x))}$. The choice of the loss function directly affects the objective function in the re-localization step. As we will show later, since sigmoid cross entropy loss is a linear function of label $y$ it leads to a {\em linear objective function} in the re-localization step. 
To speed up the re-training step, we train pairwise similarity and unary scoring functions for all the classes together by optimizing the total loss
\begin{equation}
\LL(\bpsi\mid\hat{\zb},\TT) = \sum_{c \in \CC_\TT} \LL_c(\bpsi_c, \hat{\zb}_c\mid\TT),
\end{equation}
 where $\bpsi = \left[\bpsi_c\right]_{c \in \CC_\TT}$ and $\hat{\zb} = \left[\hat{\zb}_c\right]_{c \in \CC_\TT}$ are the concatenation of respective vectors for all classes. Note that we learn the parameters of the scoring functions that minimize the loss, while $\hat{\zb}$ remains fixed in this step. Since the dataset is large, we employ Stochastic Gradient Descent (SGD) with momentum for optimization. Additionally, we subsample proposals in each bag by sampling $3$ proposals with foreground  and $7$ proposal with background label in each training iteration.

\subsection{Re-localization} \label{sec:relocalization}
In this step, we minimize the empirical loss function in \cref{eq:loss} over the feasible labeling $\hat{\zb}_c$ for the given model parameters. We first define feasible labeling set $\AA$ and simplify the objective function to an equivalent, simple linear form. Then, we discuss algorithms to optimize the objective function in the large scale settings.

\begin{figure}[t]
\begin{center}
\includegraphics[width=0.45\textwidth]{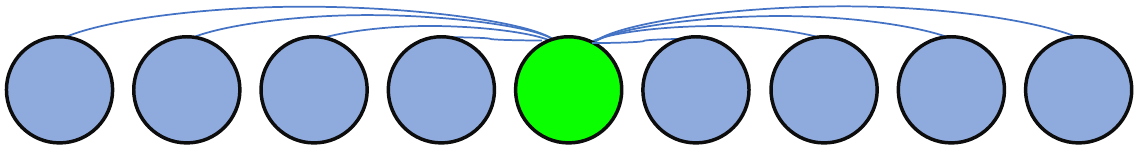}\quad\quad\quad%
\includegraphics[width=0.45\textwidth]{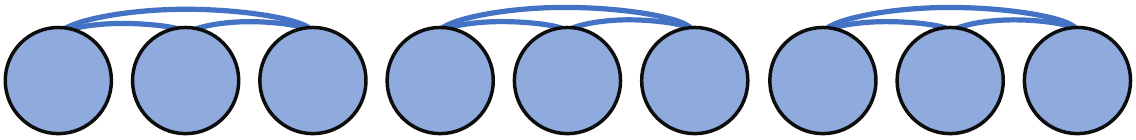}
\caption{\label{fig:ICM_update} ICM iteration (left) and initialization (right) graphical models. In both graphs, each node represents a bag (with $B$ proposals) within a dataset with $|\TT_c|=9$ bags. {\bf Left:} ICM updates the unary label of the selected node (shown in green). Edges show all the pairwise labels that gets updated in the process. Since the unary labeling of other nodes are fixed each blue edge represents $B$ elements in vector $\hat{\rb}_c$. 
{\bf Right:} For initialization we divide the dataset into smaller mini-problems (with size $K=3$ in this example) and solve each of them individually. Each edge represents $B^2$ pairwise scores that need to be computed.}
\end{center}
\end{figure}

For $\hat{\zb}_c$ to be feasible, labeling should be feasible, \ie, $\hat{\yb}_c \in \FF$ and pairwise labeling $\hat{\rb}_c$ should also be consistent with the unary labeling. For dataset $\DD_\TT$ and target class $c$, this constraint set is expressed as
\begin{equation}\label{eq:marginal_polytope}
\AA = \left\{ \begin{array}{l|l}
\multirow{3}{*}{$\hat{\zb}_c$} & \sum_{\eb\in \BB} \hat{y}_c(\eb) = Y_c(\BB)\quad \BB \in \TT\\ 
 & \sum_{\eb\in \BB} \hat{r}_c(\eb, \eb') = \hat{y}_c(\eb')\quad \BB, \BB' \in \TT, \BB' \ne \BB, \eb' \in \BB' \\
 & \hat{r}_c(\eb,\eb'), \hat{y}_c(\eb) \in \{0,1\}\quad c\in\CC, \text{for all}\ \eb\ \text{and}\ \eb'
 \end{array}
 \right\}.
\end{equation}
%
Next, we simplify the loss function in the re-localization step. Let \\
${\TT_c = \{\BB\mid \BB \in \TT, c \in \YY(\BB)\}}$ and $\TT_{\bar{c}} = \TT \setminus \TT_c$ denote the set of positive and negative bags with respect to class $c$. The loss function in \cref{eq:loss} can be decomposed into three parts
\begin{equation*}
\begin{aligned}
    \LL_c(\bpsi_c, \hat{\zb}_c|\TT) &= \LL_c(\bpsi_c, \hat{\zb}_{c}|\TT_c) + \LL_c(\bpsi_c, \hat{\zb}_{c}|\TT_{\bar{c}}) + \\
    &\sum_{\substack{\eb \in \BB \in \TT_c\\ \eb' \in \BB' \in \TT_{\bar{c}}}} \ell(\psi^\mathrm{P}_c(\eb, \eb'), \hat{r}_c(\eb, \eb')) + \ell(\psi^\mathrm{P}_c(\eb', \eb), \hat{r}_c(\eb', \eb)),
\end{aligned}
\end{equation*}
were the first two terms are the loss function in \cref{eq:loss} defined over the positive set $\TT_c$ and negative set $\TT_{\bar{c}}$, and last term is the loss defined by the pairwise similarities between these two sets. Since for any feasible labeling all the proposals in negative bags has label $0$ and remain fixed, only the value of $\LL_c(\bpsi_c, \hat{\zb}_{c}|\TT_c)$ changes within $\AA$ and other terms are constant. Furthermore, by observing that for sigmoid cross entropy loss in \cref{eq:sig_cross_entropy} we have $\ell(x,y) = \ell(x, 0) - yx$, for $y\in\left[0,1\right]$\footnote{See Appendix for the proof.}, we can further break down $\LL_c(\bpsi_c, \hat{\zb}_{c}|\TT_c)$ as
\begin{equation}\label{eq:reloc_nonvecto_form}
\begin{aligned}
    \LL_c(\bpsi_c, \hat{\zb}_{c}\mid\TT_c) &= \LL_c(\bpsi_c, \mathbf{0}\mid\TT_c) \\
    &\underbrace{-\alpha\sum_{\substack{\BB,\BB' \in \TT_c \\ \BB \neq \BB'}} \sum_{\substack{\eb \in \BB \\ \eb' \in \BB'}}  \psi^\mathrm{P}_c(\eb, \eb')\hat{r}_c(\eb, \eb') - \sum_{\BB \in \TT} \sum_{\eb \in \BB} \psi^\mathrm{U}_c(\eb) \hat{y}_c(\eb),}_{\LL_{\textrm{reloc}}(\hat{\zb}_{c}\mid\bpsi_c, \TT_c)}
\end{aligned}
\end{equation}
where $\mathbf{0}$ is zero vector of the same dimension as $\hat{\zb}_c$. Since the first term is constant with respect to $\hat{\zb}_c = \left[\hat{\yb}_c, \hat{\rb}_c\right]$, re-localization can be equivalently done by optimizing $\LL_{\textrm{reloc}}(\hat{\zb}_{c}\mid\bpsi_c, \TT_c)$ over the feasible set $\AA$
\begin{equation}\label{eq:reloc_full_opt}
\begin{gathered}
\min_{\hat{\zb}_{c}} -\alpha\hat{\rb}_c^\top\bpsi^\mathrm{P}_c - \hat{\yb}_c^\top\bpsi^\mathrm{U}_c,\\
\text{s.t. } \hat{\zb}_{c} \in \AA,
\end{gathered}
\end{equation}
where we use the equivalent vector form to represent the re-localization loss in \cref{eq:reloc_nonvecto_form}. The re-localization optimization is an Integer Linear Program (ILP) and has been widely studied in literature~\cite{schrijver1998theory}. The optimization can be equivalently expressed as a graph labeling problem with pairwise and unary potentials~\cite{savchynskyy2019discrete}. In the equivalent graph labeling problem, each bag is represented by a node in the graph where each proposal of the bag corresponds to a label of that node, and pairwise and unary potentials are equivalent to the negative pairwise similarity and negative unary scores in our problem. We discuss different graph inference methods and their limitations and present a practical method for large-scale settings. \\
%
%
%
\noindent {\bf Inference.}
Finding an optimal solution $\hat{\zb}^*_{c}$ that minimizes the loss function defined in \cref{eq:reloc_full_opt} is NP-hard and thus not feasible to compute exactly, except in small cases. Loopy belief propagation~\cite{weiss2001optimality}, TRWS~\cite{kolmogorov2006convergent}, and AStar~\cite{bergtholdt2010study}, are among the many inference algorithms used for approximate graph labeling problem. Unfortunately, finding an approximate labeling quickly becomes impractical as the size of $\TT_c$ increases, since the dimension of $\hat{\zb}_{c}$ increases quadratically with the numbers of bags in $\TT_c$ due to dense pairwise connectivity. Due to this limitation, we employ an older well-known iterated conditional modes (ICM) algorithm for optimization~\cite{besag1986statistical}. In each iteration, ICM only updates one unary label in $\hat{\yb}_c$ along with the pairwise labels that are related to this unary label while all the other elements of $\hat{\zb}_c$ are fixed. The block that gets updated in each iteration is shown in \cref{fig:ICM_update}. ICM generates monotonically non-increasing objective values and is computationally efficient. 
However, since ICM performs coordinate descent type updates and the problem in \cref{eq:reloc_full_opt} is neither convex nor differentiable as the constraint set is discrete, ICM is prone to get stuck at a local minimum and its solution significantly depends on the quality of the initial labeling.\\
\begin{center}
\scalebox{0.75}{
\begin{minipage}{1.0\linewidth}
\begin{algorithm}[H]
\DontPrintSemicolon
\KwIn{Dataset $\DD_\TT$, batch size $K$, \#epochs $E$}
\KwOut{Optimal unary labeling $\hat{\yb}^*$}
\For{$c \in \CC_\TT$}{
$T \gets round(\frac{|\TT_c|}{K})$, $\hat{\yb}_c \gets \mathbf{0}$\;
\For{$t \gets 1$ \textbf{to} $T$}{
    \tcp{Sample next mini-problem}
    $\XX \sim \TT_c$\;
    \tcp{Solve mini-problem with TRWS~\cite{kolmogorov2006convergent}}
    $\left[\bar{\yb}^*_c, \bar{\rb}^*_c\right] \gets \argmin_{\bar{\zb}_{c}}
    -\alpha\bar{\rb}_c^\top\bar{\bpsi}^\mathrm{P}_c - \bar{\yb}_c^\top\bar{\bpsi}^\mathrm{U}_c~~\text{s.t. } \bar{\zb}_{c} \in \bar{\AA}$\;
    Update corresponding block of $\hat{\yb}_c$ with $\bar{\yb}^*$
}
\tcp{Finetune for $E$ epochs}
$\hat{\yb}_c^* \gets \texttt{ICM}(\hat{\yb}_c, E)$\;
}
\Return{$\{\hat{\yb}_c^*\}_{c\in\CC_\TT}$}\;
\caption{\label{alg:relocalization} Re-localization}
\end{algorithm}
\end{minipage}%
}
\end{center}

Recent work~\cite{shaban2019learning} has shown that using accurate pairwise and unary functions learned on the source dataset, the re-localization method performs reasonably well by only looking at few bags. Motivated by this, we divide the full size problem into a set of disjoint mini-problems, solve each mini-problem efficiently using TRWS inference algorithm, and use these results to initialize the ICM algorithm.\\
\indent The initialization algorithm samples a mini-problem $\XX \in \TT_c$ and optimizes the re-localization problem $\LL_{\textrm{reloc}}(\bar{\zb}_{c}\mid\bar{\bpsi}_c, \XX)$ where vectors $\bar{\zb}_{c}$ and $\bar{\bpsi}_c$ are parts of vectors $\hat{\zb}_{c}$ and $\bpsi_c$ that are within the mini-problem defined by $\XX$ (see \cref{fig:ICM_update}). This process is repeated until all the bags in the dataset are covered. The complete re-localization step is illustrated in \cref{alg:relocalization}. \\
\indent Next, we analysis the time complexity of the re-localization step. We practically observed that computing the pairwise similarity scores is the computation bottleneck, thus we analyze the time complexities in terms of the number of pairwise similarity scores each algorithm computes. Let $M = \max_{c\in \CC_T} |\TT_c|$ denotes the maximum number of positive bags, and $B = \max_{\BB \in \TT} |\BB|$ be the maximum bag size. To solve the exact optimization in \cref{eq:reloc_full_opt}, we need to compute the vector $\bpsi_c$ with $\OO(B^2M^2)$ elements. On the other hand, each iteration of ICM only computes $\OO(BM)$ pairwise similarity scores. We additionally compute a total of $\OO(MKB^2)$ pairwise similarity scores for the initialization where $K$ is the size of the mini-problem.
Thus, ICM algorithm would be asymptotically more efficient than the exact optimization in terms of total number of pairwise similarity scores it computes, if it is run for $\Omega(MB)$ iterations or $E=\Omega(B)$ epochs. We practically observe that by initializing ICM with the result of the proposed initialization scheme it convergences in few epochs. \\
%
%
\indent Even though \cref{eq:reloc_full_opt} is similar to the DenseCRF formulation~\cite{krahenbuhl2011efficient}, the pairwise potentials are not amenable to the efficient filtering method~\cite{adams2010fast} which is the backbone of DenseCRF methods~\cite{krahenbuhl2011efficient,ajanthan2017efficient}.
Therefore, it is intractable to use any existing sophisticated optimization algorithm except for ICM and MeanField~\cite{blake2011markov}. 
Nevertheless, our block-wise application of TRWS provides an effective initialization for ICM. We additionally experimented with the block version of ICM~\cite{savchynskyy2019discrete} but it performs similarly while being slower.

\subsection{Knowledge Transfer}\label{sec:knowledge_transfer}
To transfer knowledge from the fully annotated source set $\DD_\SSX$, we first learn \emph{class generic} pairwise similarity $\psi^{\mathrm{P}}:\R^d\times\R^d\to\R$ and unary $\psi^{\mathrm{U}}:\R^d\to\R$ functions from the source set. Since the labels are available for all the proposals in the source set, learning the pairwise and unary functions is straightforward. We simply use stochastic gradient descent (SGD) to optimize the loss
\begin{equation} \label{eq:source_loss}
    \LL^T(\bpsi^{\mathrm{P}},\bpsi^{\mathrm{U}}|\SSX, \rb, \ob) = \alpha\sum_{\substack{\BB , \BB' \in \SSX \\ \BB \neq \BB'}} \sum_{\substack{\eb \in \BB \\ \eb' \in \BB'}}  \ell(\psi^{\mathrm{P}}(\eb, \eb'), r(\eb, \eb'))) + \sum_{\BB \in \SSX} \sum_{\eb \in \BB} \ell(\psi^{\mathrm{U}}(\eb), o(\eb)),
\end{equation}
where $o(\eb) \in \{0, 1\}$ is class generic objectness label, \ie,
\begin{equation}
o(\eb) = \begin{cases}
    1 \text{ if } y(\eb) \neq c_\varnothing \\
    0 \text{ otherwise,}
\end{cases}
\end{equation}
and relation function $r:\R^d\times\R^d\to\R$ is defined by \cref{eq:mil_reloc}.
Here we do not use hat notation since groundtruth proposal labels are available for the source dataset $\DD_\SSX$. We skip the details as the loss in \cref{eq:source_loss} has a similar structure to the re-training loss. Note that in general the class generic functions $\psi^{\mathrm{U}}$ and $\psi^{\mathrm{P}}$ and class specific functions $\psi_c^{\mathrm{U}}$ and $\psi_c^{\mathrm{P}}$ use different feature sets extracted from different networks. Having learned these functions, we adapt both pairwise similarity and score vectors in the re-localization step in \cref{alg:relocalization} as
\begin{gather*}
    \bpsi^{\textrm{P}}_{c} \leftarrow (1-\lambda_1) \bpsi^{\textrm{P}}_{c} + \lambda_1 \bpsi^{\textrm{P}} \\
    \bpsi^{\mathrm{U}}_c \leftarrow (1-\lambda_2) \bpsi^{\mathrm{U}}_c + \lambda_2 \bpsi^{\mathrm{U}},
\end{gather*}
where $0 \le \lambda_1, \lambda_2 \le 1$ controls the weight of transferred and adaptive functions in pairwise similarity and unary functions respectively.\\
\indent We start the alternating optimization with a \emph{warm-up} re-localization step where only the learned class generic pairwise and unary functions above are used in the re-localization algorithm, \ie, $\lambda_1,\lambda_2=1$. The warm-up re-localization step provides high quality pseudo labels to the first re-training step and speeds up the convergence of the alternating optimization algorithm.
\subsection{Network Architectures}
\label{sec:network}
{\bf\noindent Proposal and Feature Extraction} Following the experiment protocol in~\cite{uijlings2018revisiting}, we use a Faster-RCNN~\cite{ren2015faster} model trained on the source dataset $\DD_\SSX$ to extract region proposals from each image. We keep the box features in the last layer of Faster-RCNN as transferred features to be used in the class generic score functions. Following~\cite{uijlings2018revisiting,hoffman2016large,tang2016large}, we extract AlexNet~\cite{krizhevsky2012imagenet} feature vectors from each proposal as input to the class specific scoring functions $\psi^{\textrm{U}}_c$ and $\psi^{\textrm{P}}_c$.\\
{\bf\noindent Scoring Functions} Let $\eb$ and $\eb'$ denote features in $\R^d$ extracted from two image proposals. Linear layers are employed to model the class generic unary function $\psi^{\mathrm{U}}$ and all the classwise unary functions $\psi^{\mathrm{U}}_c$ i.e. $\psi^{\mathrm{U}}_c(\eb) = \wb_c^\top \eb + b_c$ where $\wb_c \in \R^d$ is the weight and $b_c \in \R$ is the bias parameter. We borrow the relation network architecture from~\cite{shaban2019learning} to model the pairwise similarity functions $\psi^{\mathrm{P}}$ and $\psi^{\mathrm{P}}_c$. The details of the relation network architecture are discussed in the Appendix.

\section{Experiments}\label{sec:experiment}
We evaluate the main applicability of our technique on different weakly supervised datasets and analyze how each part affects the final results in our method. We report the widely accepted Correct Localization (CorLoc) metric~\cite{deselaers2010localizing} for the object localization task as our evaluation metric.
\subsection{COCO 2017 Dataset}
We employ a split of COCO 2017~\cite{lin2014microsoft} dataset to evaluate the effect of different initialization strategies and our pairwise retraining and re-localization steps. The dataset has $80$ classes in total. We take the same split of~\cite{bansal2018zero,shaban2019learning} with $63$ source $\CC_\SSX$ and $17$ target $\CC_\TT$ classes and follow~\cite{shaban2019learning} to create the source and target splits to create source and target datasets with $111,085$ and $8,245$ images, respectively. 

Similar to \cite{shaban2019learning}, we use Faster-RCNN~\cite{ren2015faster} with ResNet~50~\cite{he2016deep} backbone as our proposal generator and feature extractor for knowledge transfer. We keep the top $B=100$ proposals generated by Faster-RCNN for experiments on the COCO 2017. \\
We first study different approaches for initializing the ICM method in the re-localization step. Then, we present the result of the full proposed method and compare it with other baselines.
\vspace{-0.3cm}
\subsubsection{Initialization Scheme}
\begin{figure*}[!t]
        \centering
		\includegraphics[width=0.33\linewidth]{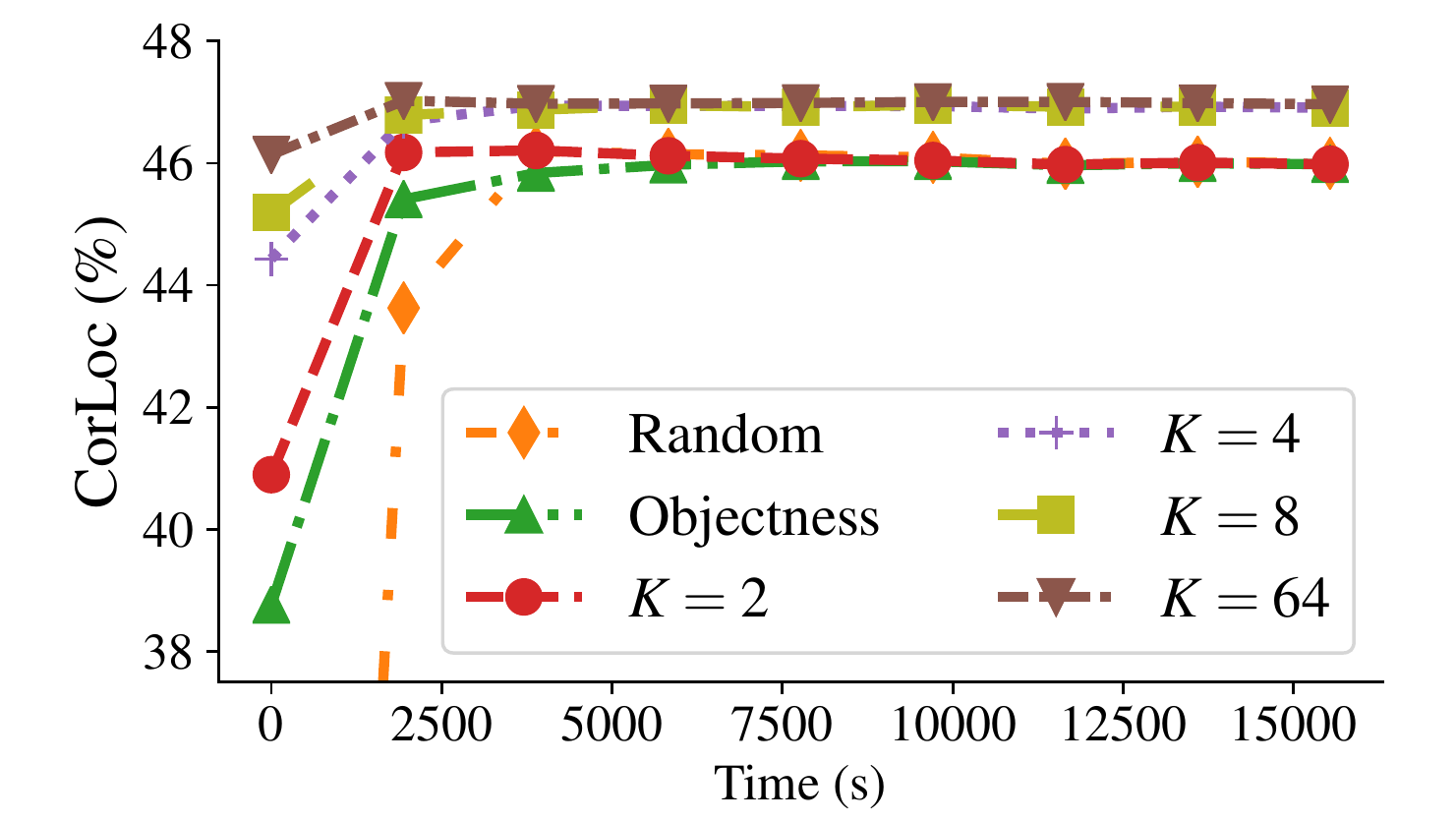} 
		\includegraphics[width=0.33\linewidth]{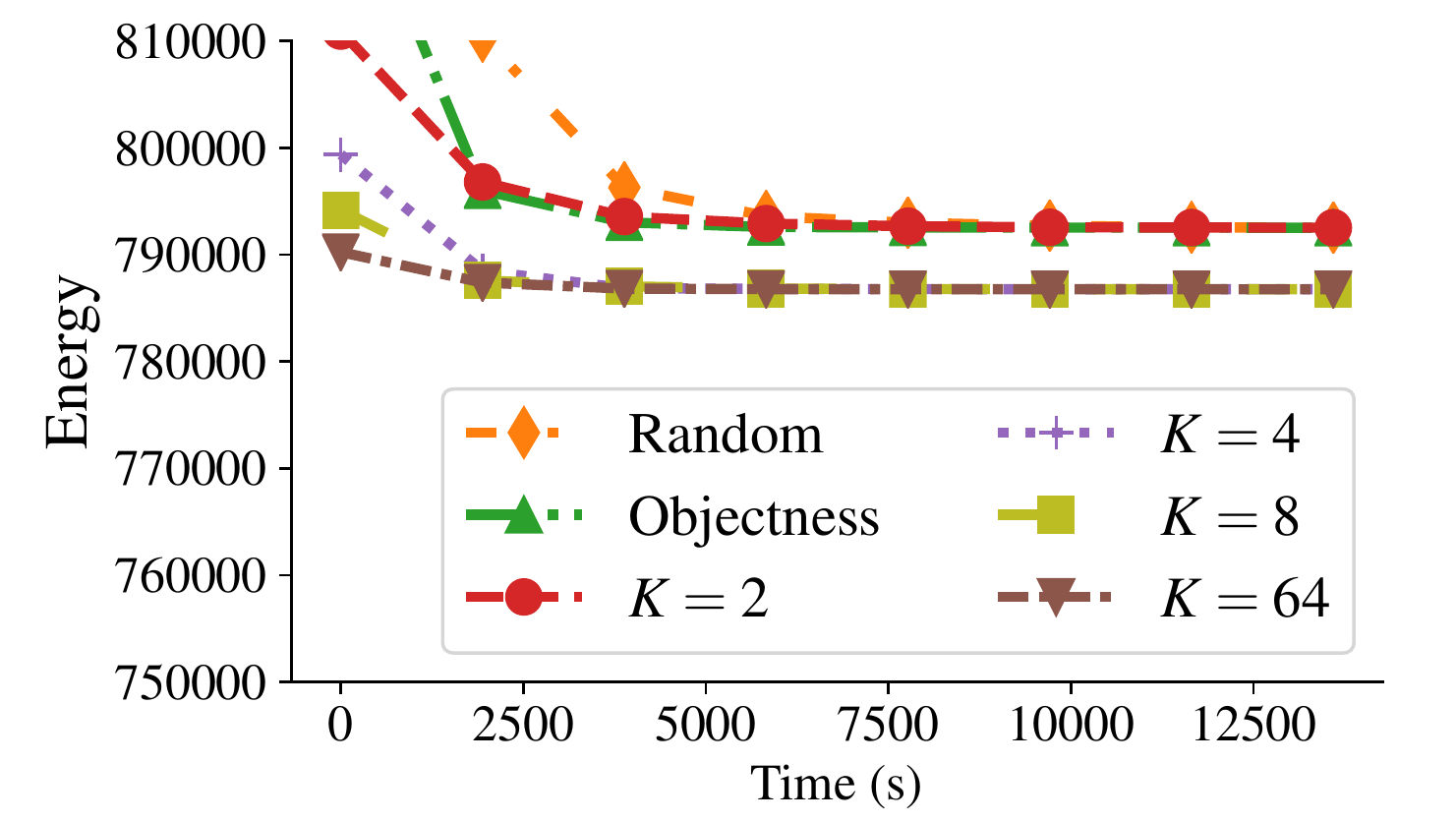}
		\includegraphics[width=0.32\linewidth]{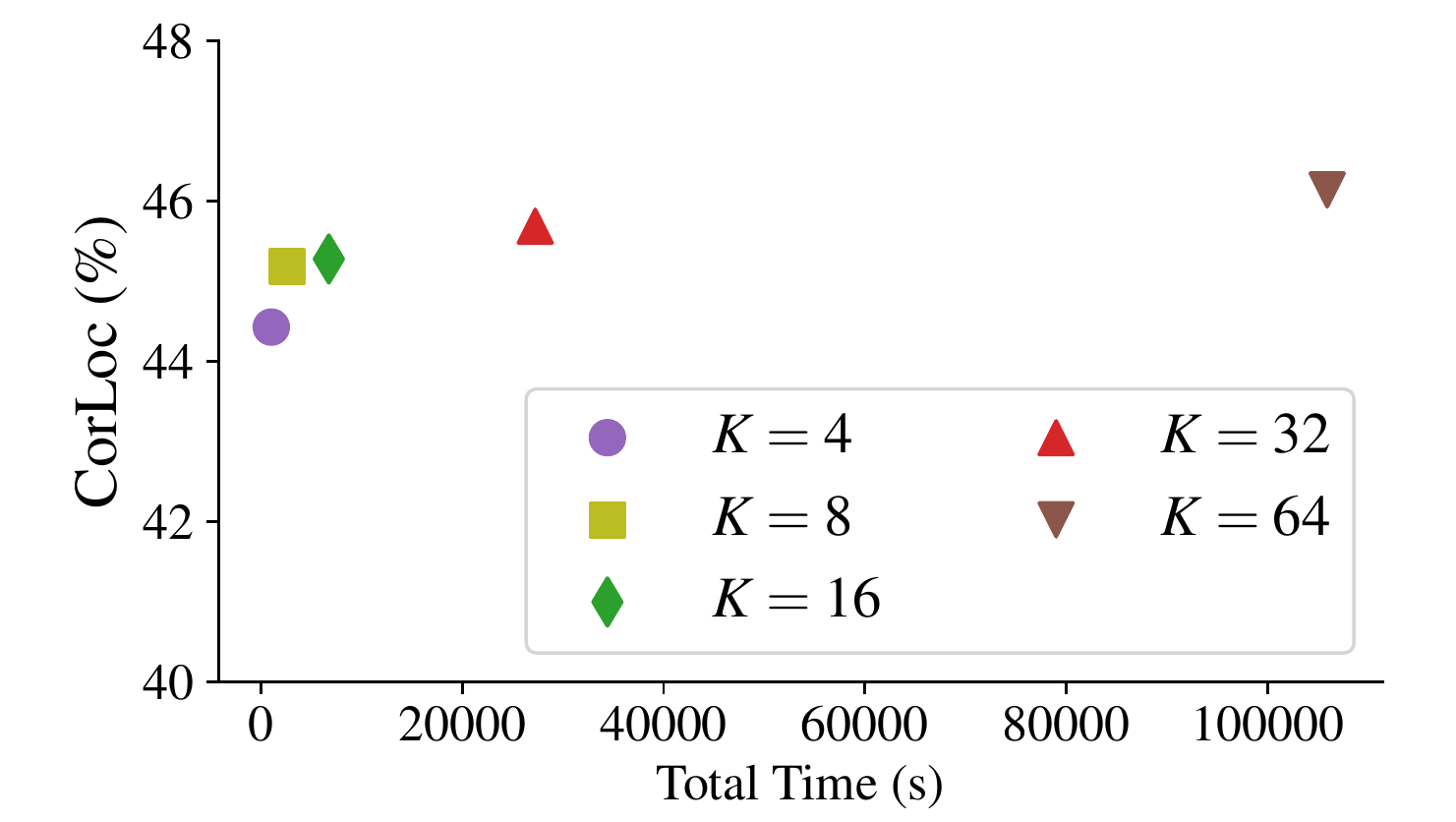}
	\vspace{-0.7cm}
	\caption{\label{fig:icm_initializaiton} {\small{ {\bf Left:} ICM CorLoc@0.5(\%) vs. time for different initialization methods. See initialization schemes for definition of each initialization method. Markers indicate start of a new epoch. ICM inference convergences in $2$ epochs and demonstrates its best performance when is initialized with the proposed initialization method. {\bf Middle:} Energy vs. time for different initialization methods. The energies in the plot are computed by summing over energies of all classes. {\bf Right:} Runtime vs. CorLoc(\%) comparison of the proposed initialization scheme with various mini-problem sizes. We observe a quadratic time increase with respect to the mini-problem size.}}}
\vspace{-0.3cm}
\end{figure*}

Since the ICM algorithm is sensitive to initialization, we devise the following experiment to evaluate different initialization methods. To limit total running time of the experiment, we only do this evaluation in the warm-up re-localization step. We start by training class generic unary and pairwise similarity scoring functions on the source dataset $\DD_\SSX$. Next, we initialize the labeling of the images in $\DD_\TT$ using the following initialization strategies:
\vspace{-0.1cm}
\begin{itemize}
    \item Random: randomly select a proposal from each bag.
    \item Objectness: select the proposal with the highest unary score from each bag.
    \item Proposed initialization method: Proposed initialization method discussed in~\cref{sec:relocalization}. We conduct the experiment with different mini-problem sizes $K \in \{2,4,8,64\}$. We use TRWS~\cite{kolmogorov2006convergent} algorithm for inference in each mini-problem.
\end{itemize}
\vspace{-0.1cm}
Finally, we perform ICM with each of the initialization methods. \cref{fig:icm_initializaiton} shows the CorLoc and Energy vs. time plots as well as the computation time for different initialization methods. As expected, $K=64$ exhibits the best initialization performance. However, ICM converges to similar energy when $4 \leq K \leq 64$ is used in the initialization method. In the extreme case with mini-problem of size $K=2$, ICM converges to a worse local minimum in terms of CorLoc and energy value. Surprisingly, random initialization converges to the same result as objectness and $K=2$. We also tried initializing ICM with the proposal that covers the complete image as it is the initialization scheme that is commonly used in MIL alternating optimization algorithms~\cite{uijlings2018revisiting,cinbis2016weakly}. But it performs significantly worse  than the other initialization methods.

%
These results highlight the importance of initialization in ICM inference. 
Note that increasing $K$ beyond $64$ might provide a better initialization to ICM and increase the results further but it quickly becomes impractical as the time plot in \cref{fig:icm_initializaiton} illustrates. As a rule of thumb, one should increase the mini-problem size as far as time and computational resources allow.
\vspace{-0.3cm}
\subsubsection{Full Pipeline} 
Here, we conduct an experiment to determine the importance of learning pairwise similarities on the COCO dataset. We compare our full method with the unary method which only learns and uses unary scoring functions during, warm-up, re-training and re-localization steps. This method is analogous to \cite{uijlings2018revisiting}. The difference is that it uses cross entropy loss and SGD training instead of Support Vector Machine used in~\cite{uijlings2018revisiting}. Also, we do not employ hard-negative mining after each re-training step. For this experiment, we use mini-problems of size $K=4$ for initializing ICM. We run both methods for $5$ iterations of alternating optimization on the target dataset. Our method achieves ${\bf 48.3\%}$ compared to $39.4\%$ CorLoc@0.5 of the unary method. This clearly shows the effectiveness of our pairwise similarity learning.


\vspace{-0.3cm}
\subsection{ILSVRC 2013 Detection Dataset}
We closely follow the experimental protocol of~\cite{uijlings2018revisiting,hoffman2016large,tang2016large} to create source and target datasets on ILSVRC 2013~\cite{russakovsky2015imagenet} detection dataset. The dataset has $200$ categories with full bounding box annotations. We use the first $100$ alphabetically ordered classes as source categories $\CC_\SSX$ and the remaining $100$ classes as target categories $\CC_\TT$. 
%
The dataset is divided into source training set $\DD_\SSX$ with $63$k images, target set $\DD_\TT$ with $65$k images, and a target test set with $10$k images. We report CorLoc of different algorithms on $\DD_\TT$. Similar to previous works~\cite{uijlings2018revisiting,hoffman2016large,tang2016large}, we additionally train a detector from the output of our method on target set $\DD_\TT$, and evaluate it on the target test set.

For a fair comparison, we use a similar proposal generator and multi-fold strategy as~\cite{uijlings2018revisiting}. We use Faster-RCNN~\cite{ren2015faster} with Inception-Resnet~\cite{szegedy2017inception} backbone trained on source dataset $\DD_\SSX$ for object bounding box generation. 
The experiment on COCO suggests a small mini-problem size $K$ would be sufficient to achieve good performance in the re-localization step. We use $K=8$ to balance the time and accuracy in this experiment.

\begin{table}[t]
\begin{center}
\caption{\label{table:ilsvrc13}{\small{Performance of different methods on ILSVRC 2013. Proposal generators and their backbone models are shown in the second and third column. Total time is shown in ``Training+Inference'' format. CorLoc is reported on the target set. The last column shows the performance of an object detector trained on the target set and evaluated on the target test set.  $^*$The first $3$ methods use RCNN detector with AlexNet backbone while other methods utilize Faster-RCNN detector with Inception-Resnet backbone.}}}
\resizebox{1.0\linewidth}{!}{%
\begin{tabular}{l c c c  c  c || c}
\toprule
Method & Proposal Generator & Backbone & CorLoc@0.5 & CorLoc@0.7 & Time(hours) & mAP@0.5 \\
\midrule
LSDA \cite{hoffman2016large} & Selective Search~\cite{uijlings2013selective} & AlexNet~\cite{krizhevsky2012imagenet} & 28.8 & - & - & $18.1^*$\\
Tang~\etal~\cite{tang2016large} & Selective Search~\cite{uijlings2013selective} & AlexNet~\cite{krizhevsky2012imagenet} & - & - & - & $20.0^*$\\
Uijlings \etal\cite{uijlings2018revisiting} & SSD~\cite{liu2016ssd} & Inception-V3~\cite{szegedy2016rethinking} & 70.3 & 58.8 & - & $23.3^*$ \\
Uijlings \etal\cite{uijlings2018revisiting} & Faster-RCNN & Inception-Resnet & 74.2 & 61.7 & - & $36.9^\text{ }$ \\
\midrule
Warm-up (unary) & Faster-RCNN & Inception-Resnet & 68.9 & 59.5 & 0 & - \\
Warm-up & Faster-RCNN & Inception-Resnet & 73.8 & 62.3 & 5+3 & -\\
Unary & Faster-RCNN & Inception-Resnet & 72.8 & 62.0 & 13+2 & $38.1^\text{ }$ \\
Full (ours) & Faster-RCNN & Inception-Resnet & {\bf 78.2} & {\bf 65.5}  & 65+13 & $\mathbf{41.7}^\text{ }$\\
\midrule
Supervised\cite{uijlings2018revisiting} & & &  & &  & 46.2\\
\bottomrule
\end{tabular}
} 
\end{center}
\vspace{-.5cm}
\end{table}

\subsubsection{Baselines and Results} 
\begin{figure}[t]
\begin{center}
\includegraphics[width=0.98\textwidth]{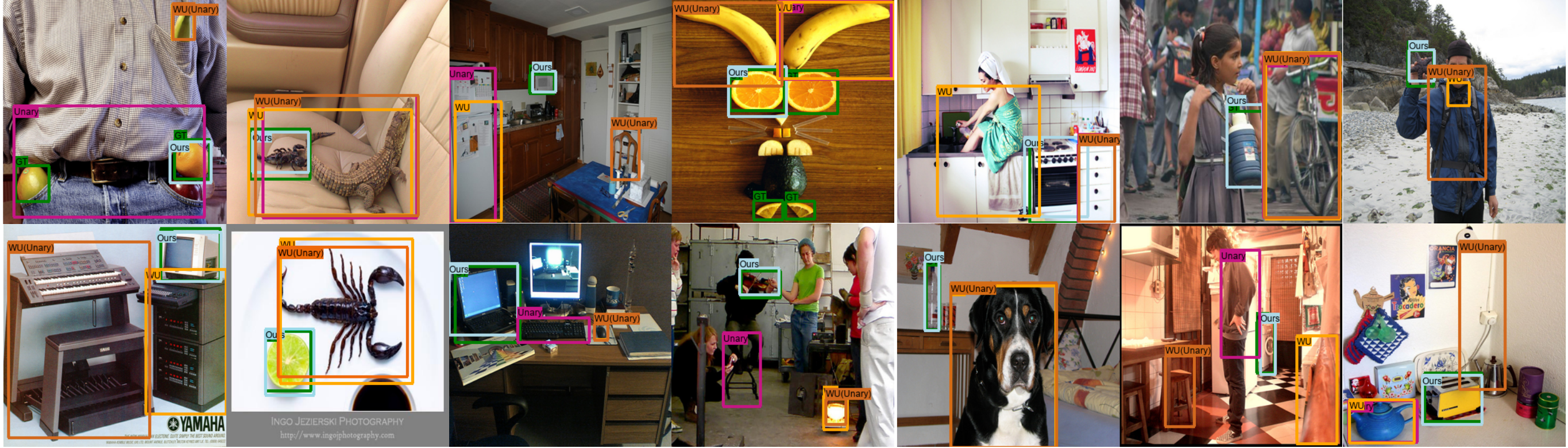}
\vspace{-0.1cm}
\caption{\label{fig:qualitative_results} {Success cases on ILSVRC 2013 dataset. Unary method that relies on the objectness function tends to select objects from source classes that have been seen during training. Note that ``banana'', ``dog'', and ``chair'' are samples from source classes. Bounding boxes are tagged with method names. ``GT'' and ``WU'' stand for groundtruth and warm-up respectively. See Appendix for a larger set of success and failure cases.}}
\end{center}
\vspace{-0.5cm} 
\end{figure}

We compare our method with two knowledge transfer techniques\cite{hoffman2016large,uijlings2018revisiting} for WSOL. In addition, we demonstrate the results of the following baselines that only use unary scoring function: \begin{itemize}
    \item Warm-up (unary): To see the importance of learning pairwise similarities in knowledge transfer, we perform the warm-up re-localization with only the transferred unary scores $\bpsi^{\mathrm{U}}$. This can be achieved by simply selecting the box with the highest unary score within each bag. We compare this results with the result of the warm-up step which uses both pairwise and unary scores in knowledge transfer.
    \item Unary: Standard MIL objective in \cref{eq:rev_opt} which only learns labeling and the unary scoring function. 
\end{itemize}
We compare these results with our full pipeline which starts with a warm-up re-localization step followed by alternating re-training and re-localization steps. 
The results are illustrated in \cref{table:ilsvrc13}. Compared to Uijlings \etal\cite{uijlings2018revisiting}, our method improves the CorLoc@0.5 performance on the target set by $4\%$ and mAP@0.5 on the target test set by $4.8\%$. Warm-up re-localization improves CorLoc performance of warm-up (unary) by $4.9\%$ with transferring a pairwise similarity measure from the source classes. Note that the result of warm-up step without any re-training performs on par with the Uijlings \etal\cite{uijlings2018revisiting} MIL method. The CorLoc performance at the stricter {\texttt{IoU}}$>0.7$ also shows similar results. Some of the success cases are shown in \cref{fig:qualitative_results}. \\
Compared to~\cite{uijlings2018revisiting}, our implementation of the MIL method performs worse with {\texttt IoU} threshold $0.5$ but better with stricter threshold $0.7$. We believe the reason is having a different loss function and hard-negative mining in~\cite{uijlings2018revisiting}.
\section{Conclusion}
We study the problem of learning localization models on target classes from weakly supervised training images, helped by a fully annotated source dataset. 
We adapt MIL localization model by adding a classwise pairwise similarity module that learns to directly compare two input proposals. 
Similar to the standard MIL approach, we learn the augmented localization model and annotations jointly by  two-step alternating optimization. We represent the re-localization step as a graph labeling problem and propose a computationally efficient inference algorithm for optimization.
Compared to the previous work~\cite{deselaers2010localizing} that uses pairwise similarities for this task, the proposed method is represented in alternating optimization framework with convergence guarantee and is computationally efficient in large-scale settings. 
The experiments show that learning pairwise similarity function improves the performance of WSOL over the standard MIL.
\subsection*{Acknowledgement}
\footnotesize{We would like to thank Jasper Uijlings for providing helpful instructions and their original models and datasets. We gratefully express our gratitude to Judy Hoffman for her advice on improving the experiments. We also thank the anonymous reviewers for their helpful comments to improve the paper.
This research is supported in part by the Australia Research Council Centre of Excellence for Robotics Vision
(CE140100016).}
\clearpage
\bibliographystyle{splncs}
\bibliography{egbib}

\clearpage
\appendix
\begin{center}
     {\bf \Large Pairwise Similarity Knowledge Transfer  for\\
     Weakly Supervised Object Localization \\ Supplementary Material}\\ 
     
 \end{center}

\section{Missing Proof}
Let $\ell:\R\times\R\to\R$ be the sigmoid cross-entropy loss function 
\[
\ell(x, y) = -(1-y) \log(1-\sigma(x)) - y \log(\sigma(x)),
\]
where ${\sigma(x)=1/(1+\exp(-x))}$ is the sigmoid function. Then, $\ell(x,y) = \ell(x, 0) - yx$, for any $x\in\R$ and $y\in\left[0,1\right]$.
\begin{proof}
\begin{align*}
    \ell(x, y) - \ell(x, 0) &= \Big(-(1-y) \log(1-\sigma(x)) - y \log(\sigma(x))\Big) + \log(1-\sigma(x))\\
    &= y \log(1-\sigma(x)) - y \log(\sigma(x)) \\
    &= -yx
\end{align*}
Last equality is derived using the fact that ${\log(1-\sigma(x)) - \log(\sigma(x)) = -x}$ which can be easily verified by plugging in the sigmoid function.
\end{proof}

\section{Experiment Details}
The CorLoc measure used to evaluate the performance of different WSOL methods is the ratio of correctly localized objects in each class and computes the mean over all classes. A localization is correct if it has Intersection-over-Union~({\texttt IoU}) greater than a threshold with the groundtruth object bounding box. We report with $0.5$ and $0.7$ thresholds in our experiments. All experiments are done on a single Nvidia GTX 1080 GPU and 3.2GHz Intel(R) Xeon(R) CPU with 128 GB of RAM. 

\subsection{Relation Network Architecture} The relation network~\cite{sung2018learning} $s:\R^d \times \R^d\to\R$ has two modules. We use the same architecture proposed in~\cite{shaban2019learning} in our method. First module maps both input features into a joint feature space using embedding function $\EE: \R^d \times \R^d \rightarrow \R^d$ and is defined as
\[
    \EE(\eb, \eb') =  \tanh(W_1 \left[\eb, \eb'\right] + \bb_1)  \sigma(W_2 \left[\eb, \eb'\right] + \bb_2) 
    + \frac{\eb + \eb'}{2},
\]
where $W_1, W_2 \in \R^{d\times 2d}$ and vectors $\bb_1, \bb_2 \in \R^{d}$ are the parameters of the feature embedding module and $\tanh$ and $\sigma$ are hyperbolic tangent and sigmoid activation functions respectively. Finally, a linear layer maps these features into similarity score
\[
s(\eb, \eb') = \wb^\top\EE(\eb, \eb') + b,
\]
where $\wb \in\R^d$ and $b \in \R$. We share the parameters of the embedding functions in $\psi^{\mathrm{P}}_c(\eb, \eb')$ for all the classes $c \in \CC_\TT$ to reduce the number of parameters.
\subsection{ILSVRC 2013 Detection Experiment} We use an identical dataset and bounding box proposals as were used in~\cite{uijlings2018revisiting}\footnote{Thanks to Jasper Uijlings for sharing their pre-trained Faster-RCNN model and image splits}. 
The dataset is created by following the dataset generation protocol of~\cite{hoffman2016large,tang2016large}. The val1 split is augmented with images from the training set such that each class has $1000$ annotated bounding boxes in total~\cite{girshick2014rich}. The dataset has $200$ categories with full bounding box annotations. The first $100$ alphabetically ordered classes are used as source categories $\CC_\SSX$ and the remaining $100$ classes are used as target categories $\CC_\TT$. The source dataset $\DD_\SSX$ is formed by all images in the augmented val1 set that have an object in $\CC_\SSX$. As for our target dataset $\DD_\TT$, all images which have an object in the target categories $\CC_\TT$ are used and all the bounding box annotations are removed and only the bag labels $\YY_\TT$ are kept. The dataset has $63$k and $65$k images in source and target sets, respectively.

We also use the same Faster-RCNN~\cite{ren2015faster} model with Inception-Resnet~\cite{szegedy2017inception} backbone trained on the source dataset $\DD_\SSX$ that were used in~\cite{uijlings2018revisiting} for object bounding box generation. Since the Faster-RCNN model used in~\cite{uijlings2018revisiting} is trained with a class agnostic loss and the feature vectors are not suitable for discriminating between different classes we also trained the same architecture with a multi-class loss and use it to extract the features from the bounding box proposals of~\cite{uijlings2018revisiting}.

To have a fair comparison to~\cite{uijlings2018revisiting}, we perform multi-folding strategy~\cite{cinbis2016weakly} to avoid overfitting: the target dataset is split into $10$ random folds and then re-training is done on $9$ folds while re-localization is performed on the remaining fold. Values of hyper-parameters are obtained using cross validation. 

\section{More Qualitative Results}
\begin{figure}[!h]
\begin{center}
\includegraphics[width=0.98\textwidth]{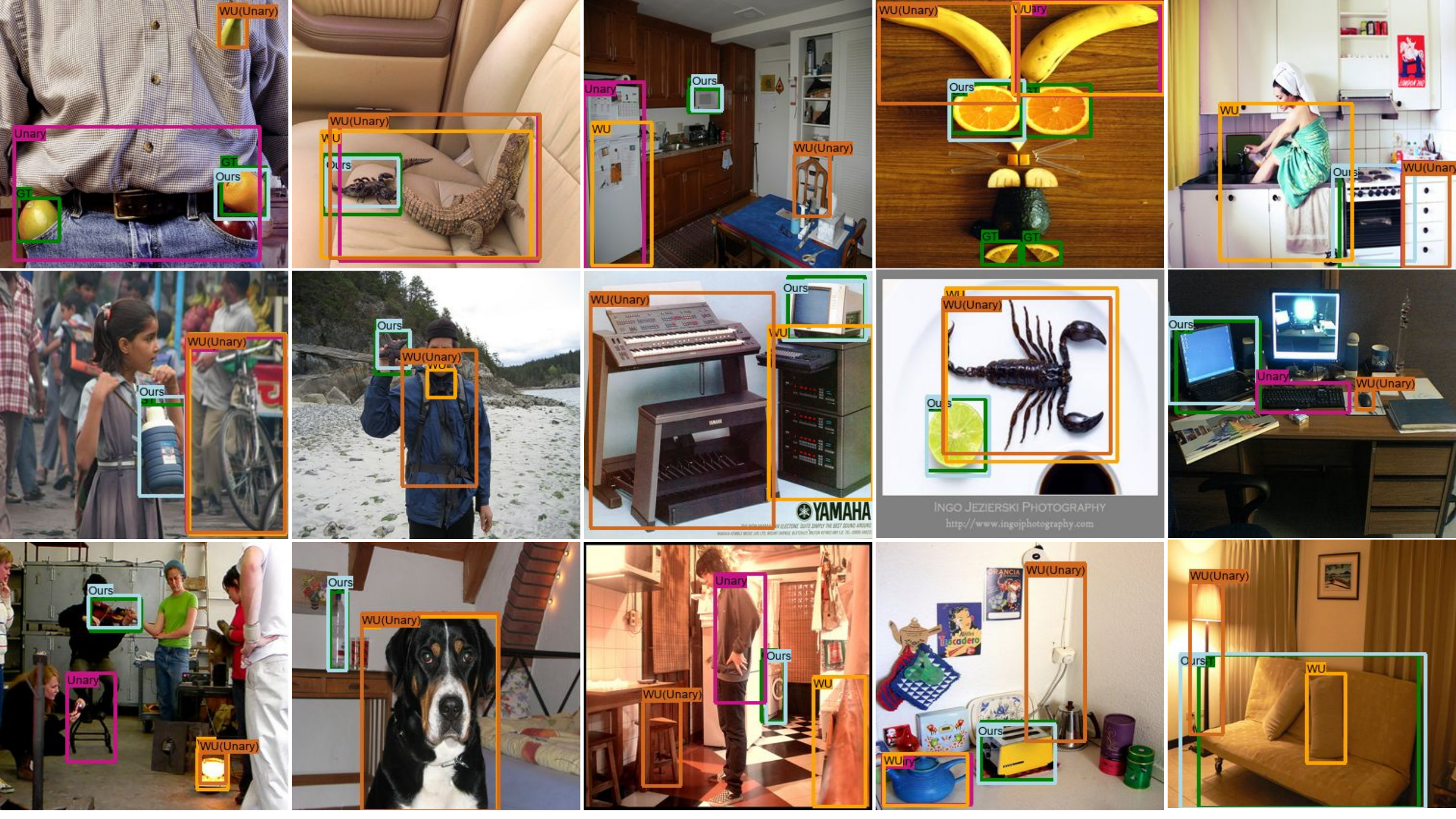} \\
\vspace{-0.08cm}
\hspace{0.01cm}
\includegraphics[width=0.98\textwidth]{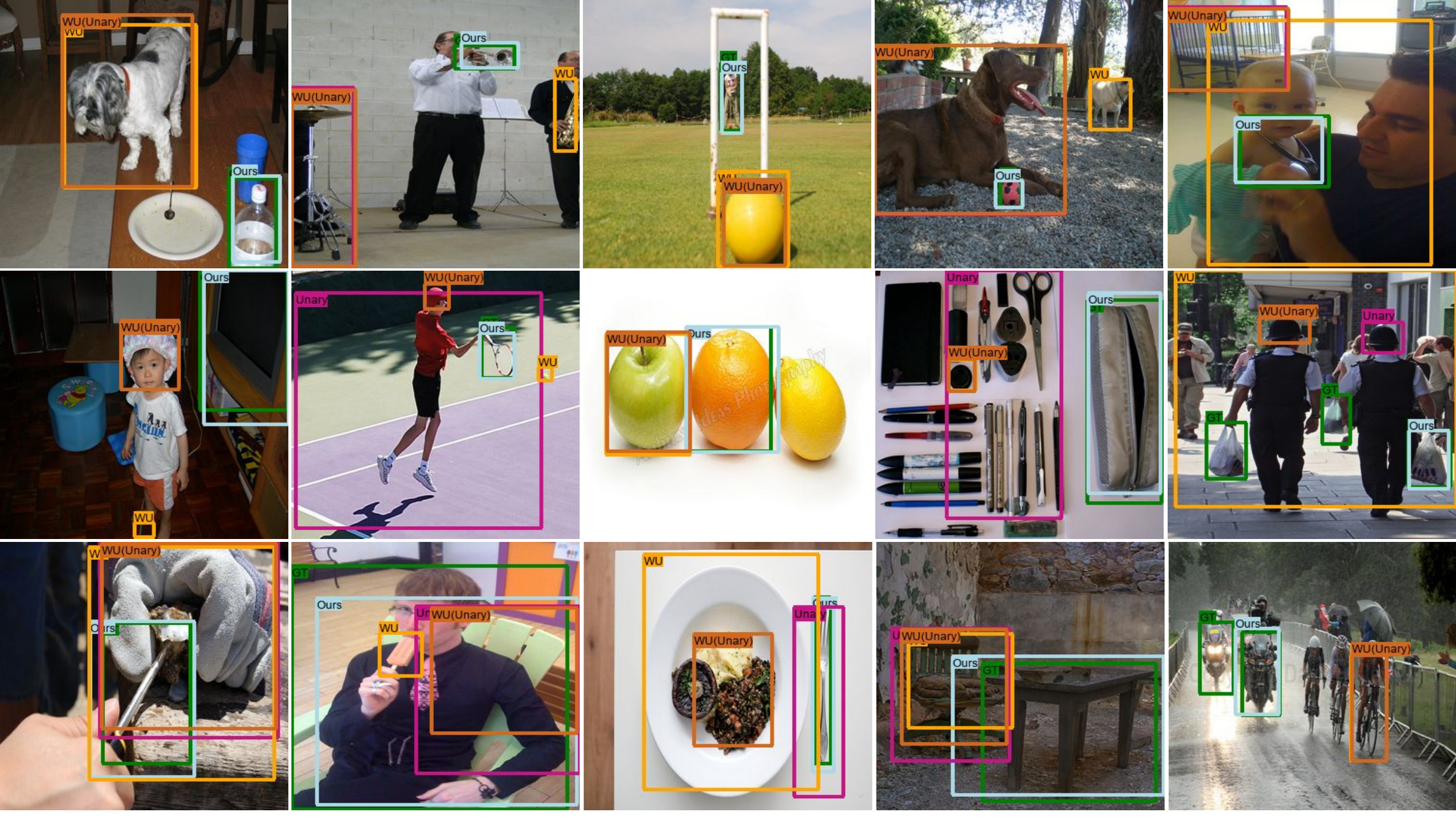}
\vspace{-0.1cm}
\caption{\label{fig:high_qual} {Extended results of~\cref{fig:qualitative_results}}}
\end{center}
\vspace{-0.5cm} 
\end{figure}

\begin{figure}[!h]
\begin{center}
\includegraphics[width=0.98\textwidth]{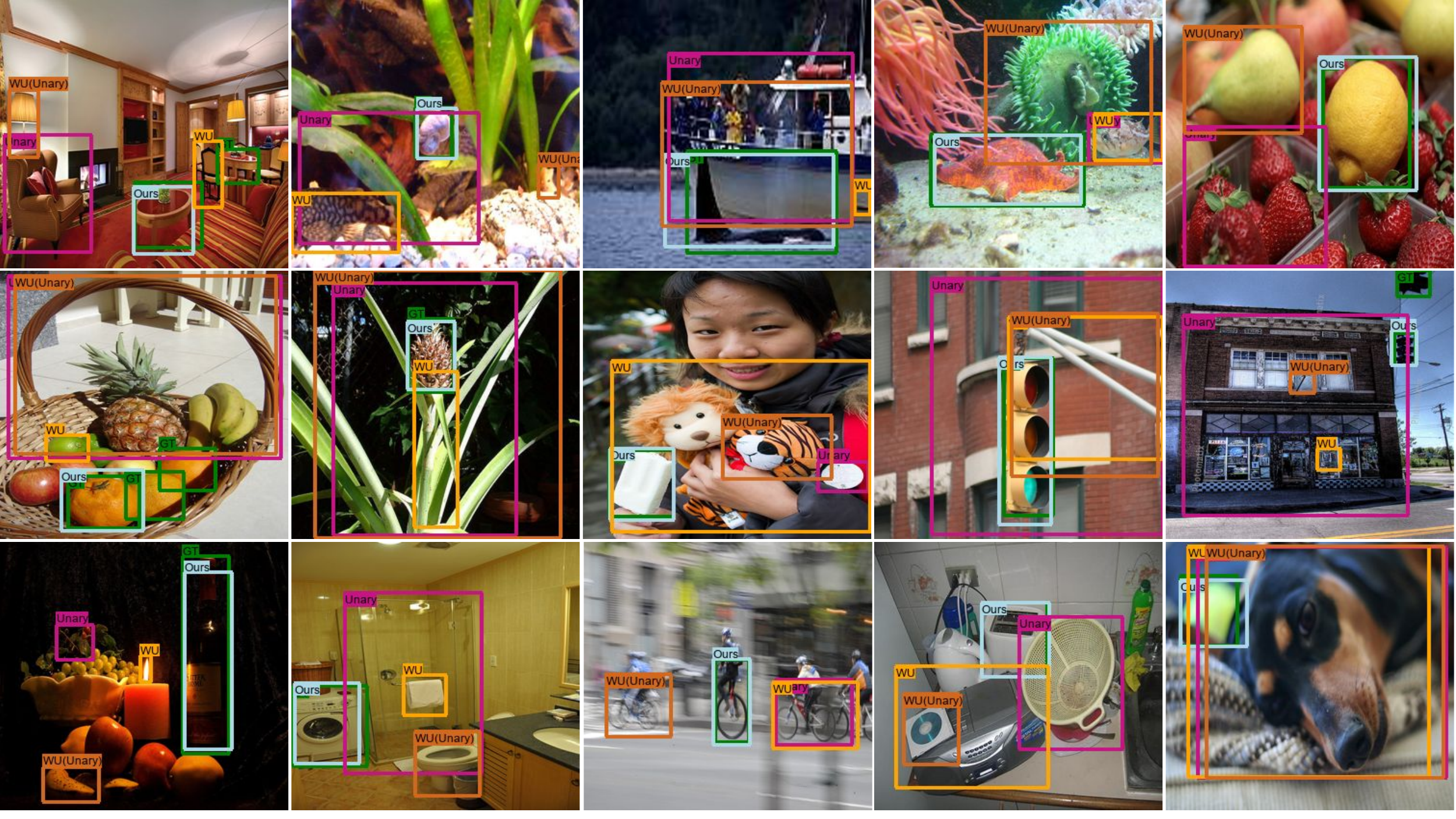} \\
\vspace{-0.08cm}
\hspace{0.01cm}
\includegraphics[width=0.98\textwidth]{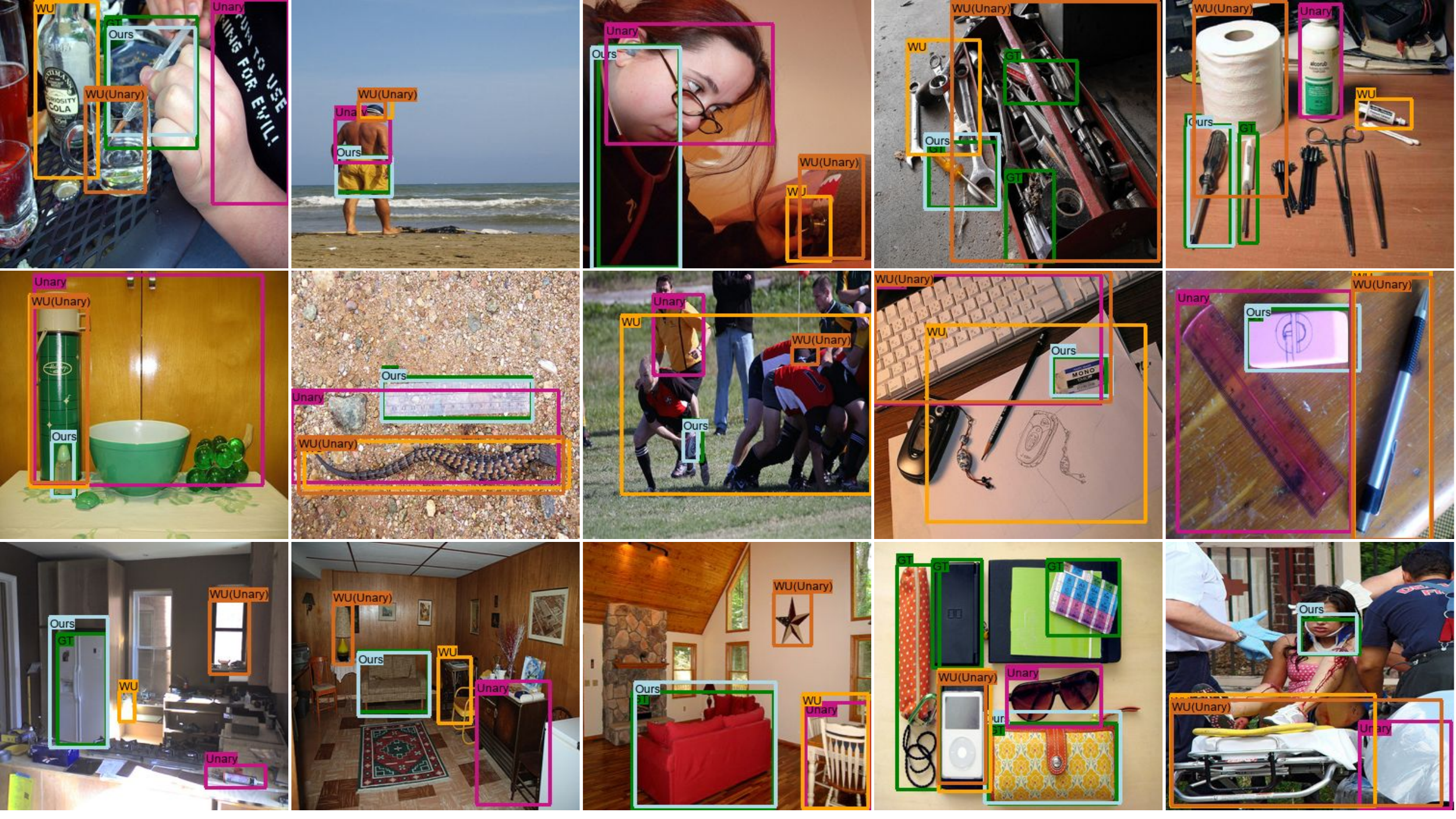}
\vspace{-0.1cm}
\caption{\label{fig:page3} {Extended results of~\cref{fig:qualitative_results}}}
\end{center}
\vspace{-0.5cm} 
\end{figure}

\begin{figure}[!h]
\begin{center}
\includegraphics[width=0.98\textwidth]{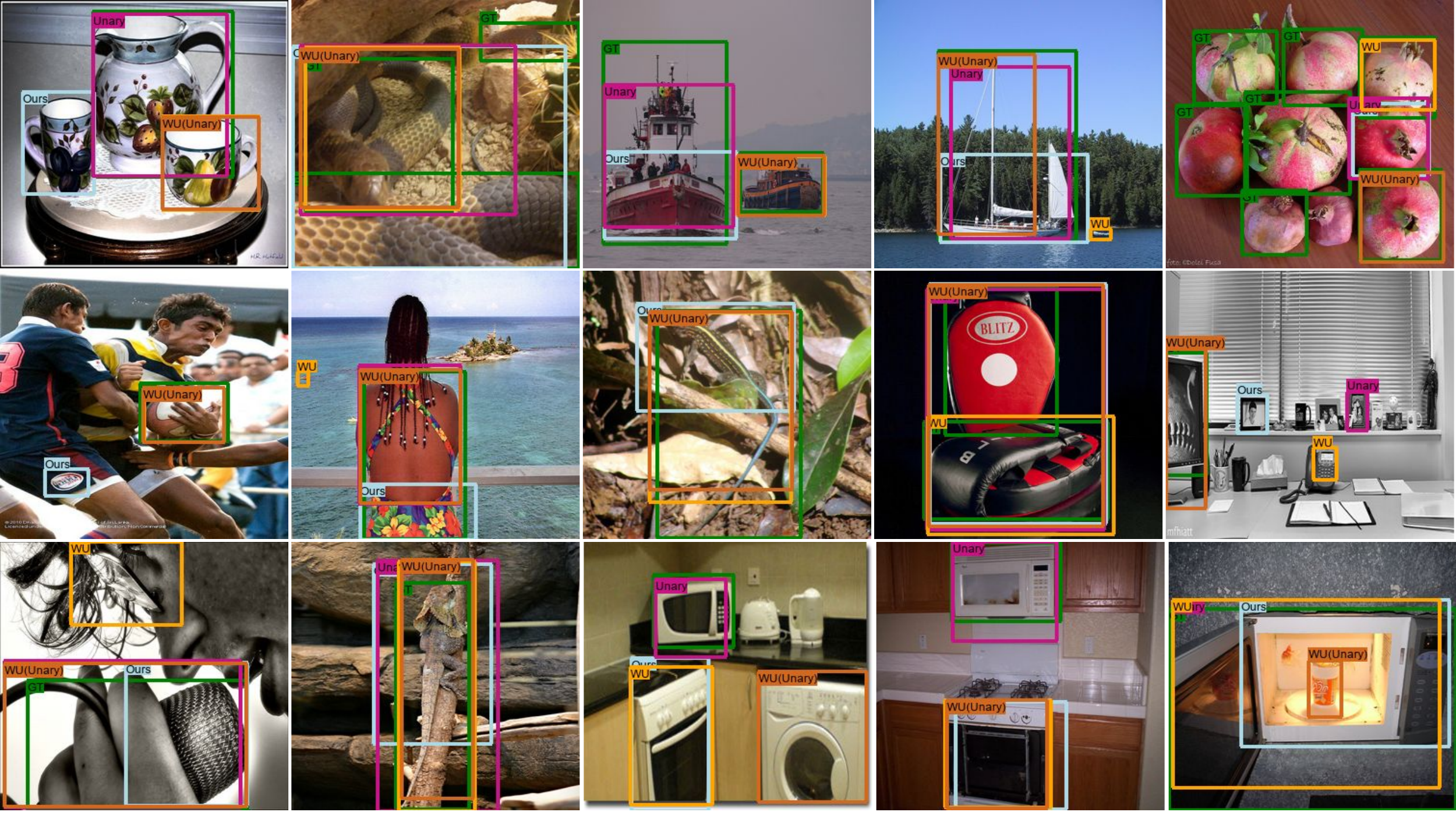}
\caption{\label{fig:failure} {Failure cases on ILSVRC 2013 dataset.}}
\end{center}
\vspace{-0.5cm} 
\end{figure}

\begin{figure}[!h]
\begin{center}
\includegraphics[width=0.98\textwidth]{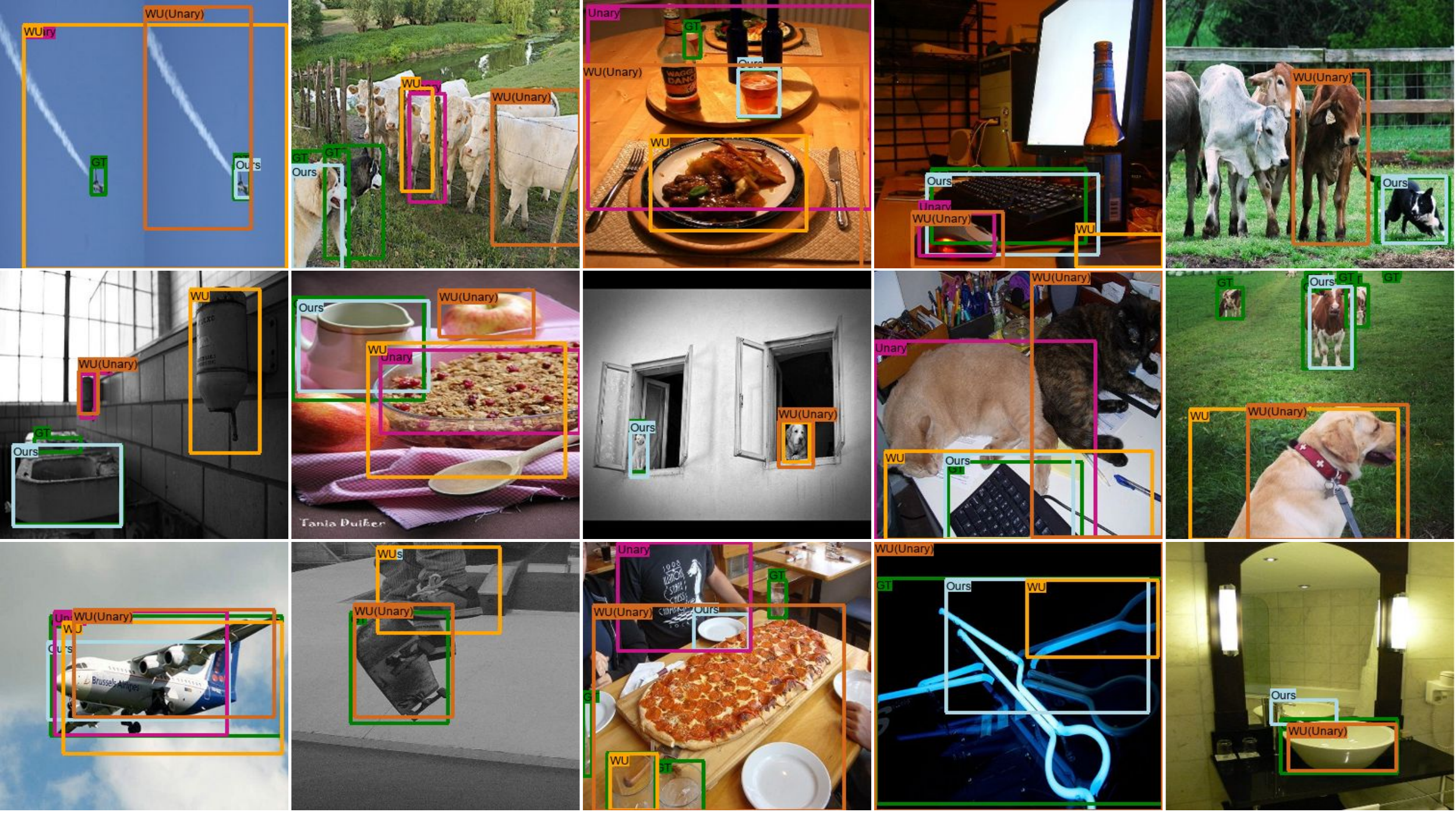}
\caption{\label{fig:coco} {Success and failure cases on COCO dataset. First two rows show the success cases of our method while the last row shows the failure cases.}}
\end{center}
\vspace{-0.5cm} 
\end{figure}

Qualitative results on ILSVRC 2013 dataset are illustrated in~\cref{fig:high_qual}, and~\cref{fig:page3}. Failure cases on this dataset are also presented in~\cref{fig:failure}. Refer to \cref{fig:qualitative_results} for more information on bounding box tags. Overall, selection of a visually similar object in the image, occlusion and disconnected objects, multi-part objects, and even errors in dataset annotations are the source of most of the failures on this dataset. \cref{fig:coco} shows the qualitative results on the COCO dataset.

\end{document}